\documentclass[10pt]{article}

\usepackage[a4paper,margin=1in]{geometry}
\usepackage{microtype}

\usepackage{amsmath,amssymb}

\usepackage{booktabs}
\usepackage{multirow}
\usepackage{graphicx}
\usepackage{subcaption}
\usepackage{authblk}

\graphicspath{{figs/}}

\usepackage[numbers]{natbib}

\usepackage[hidelinks]{hyperref}

\title{Rotation-Robust Regression with Convolutional Model Trees}

\author[1]{Hongyi Li}
\author[2]{William Ward Armstrong}
\author[1]{Jun Xu}

\affil[1]{Department of Automation, Harbin Institute of Technology, Shenzhen, China}
\affil[2]{Department of Computing Science, University of Alberta, Edmonton, Canada}

\date{}

\begin{document}
\maketitle

\begin{abstract}
We study rotation-robust learning for image inputs using \emph{Convolutional Model Trees} (CMTs)~\cite{CMT_original}, whose split and leaf coefficients can be structured on the image grid and transformed geometrically at deployment time.
In a controlled MNIST setting with a rotation-invariant regression target, we introduce three geometry-aware inductive biases for split directions---convolutional smoothing, a tilt dominance constraint, and importance-based pruning---and quantify their impact on robustness under in-plane rotations.
We further evaluate a \emph{deployment-time orientation search} that selects a discrete rotation maximizing a forest-level confidence proxy without updating model parameters.
Orientation search improves robustness under severe rotations but can be harmful near the canonical orientation when confidence is misaligned with correctness.
Finally, we observe consistent trends on MNIST digit recognition implemented as one-vs-rest regression, highlighting both the promise and limitations of confidence-based orientation selection for model-tree ensembles.
\end{abstract}

\section{Introduction}
Real-world vision systems frequently encounter inputs under nuisance transformations such as unknown in-plane rotation.
When the prediction target is transformation-invariant, a desirable learner should maintain stable performance across orientations rather than relying on a single canonical pose.
While invariance can often be induced via architectural design or augmentation, it remains less straightforward for model-tree ensembles with linear components, which are attractive in low-resource and interpretable settings.
This paper revisits rotation robustness through \emph{Convolutional Model Trees} (CMT s)~\cite{CMT_original}, where split and leaf coefficients are explicitly organized on the image grid and can be transformed geometrically at deployment time.

We study rotation-robust regression on image inputs using a model-tree ensemble with geometry-aware split constraints.
We choose a task where the label is rotation-invariant by construction: predicting the discrete perimeter of a digit silhouette extracted from MNIST images.
A single model is trained once and evaluated on inputs rotated by angles from $-60^\circ$ to $+60^\circ$; a robust model should maintain similar prediction accuracy across this range.

\paragraph{Deployment-time orientation search (OS).}
Beyond training-time robustness, we study a deployment-time procedure that changes \emph{only} the input orientation used for prediction while keeping all model parameters fixed.
Given a small discrete set of candidate rotations, OS selects the orientation that maximizes a margin-based confidence score aggregated along the forest's decision paths, and then predicts using the forest under that selected orientation.
Although we evaluate this procedure on rotated inputs, it is intended for deployment-time use when the input orientation is unknown after the model has been trained and validated.

Our contributions are:
\begin{itemize}
  \item We implement three geometry-aware inductive biases for CMT split directions: convolutional smoothing, a tilt dominance constraint motivated by the CMT analysis~\cite{CMT_original}, and importance-based pruning.
  \item We evaluate a deployment-time OS procedure for discrete orientation selection based on a forest-level confidence proxy, without retraining.
  \item We provide a full ablation on MNIST and analyze both predictive performance and compute cost, clarifying when orientation selection helps and when it hurts.
\end{itemize}

\section{Method}
\subsection{Notation}
An input grayscale image is $I \in [0,1]^{H\times W}$ with $H=W=28$.
We vectorize it to $x\in\mathbb{R}^{D}$ where $D=HW$.
At a tree node, let $X\in\mathbb{R}^{n\times D}$ denote the node's design matrix and $y\in\mathbb{R}^n$ the regression targets.

\subsection{Task Definition: Perimeter Regression}
Given an image $I$, we define a binary mask $M = \mathbb{I}[I > \tau]$ with threshold $\tau=0.1$.
The regression target is the discrete perimeter computed from horizontal and vertical boundary changes:
\begin{equation}
y = \alpha \Big( \sum_{u,v} \mathbb{I}[M_{u,v} \neq M_{u-1,v}] + \sum_{u,v} \mathbb{I}[M_{u,v} \neq M_{u,v-1}] \Big),
\end{equation}
with scaling $\alpha=1/100$.
This definition counts both outer boundaries and interior holes, since each hole boundary induces additional mask disagreements.
The resulting target is approximately invariant to in-plane rotations of the input.

\subsection{Midpoint Hyperplane Split}
At each node, the samples induce an axis-aligned bounding box in pixel space.
Let $x_{\min}, x_{\max} \in \mathbb{R}^D$ denote per-feature bounds of the node data; define the box midpoint and half-width:
\begin{equation}
m = \tfrac{1}{2}(x_{\min}+x_{\max}), \quad h = \tfrac{1}{2}(x_{\max}-x_{\min}).
\end{equation}
Given a split normal $w \in \mathbb{R}^{D}$, we compute the signed routing score
\begin{equation}
g(x) = (x - m)^\top w,
\end{equation}
and route the sample to the left child if $g(x)\le 0$ and to the right child otherwise.

\subsection{Leaf Models: Least-Squares Direction (Ridge-Optional) via Adam}
We fit a (possibly ridge-regularized) linear model at each node:
\begin{equation}
\min_{w} \ \frac{1}{2n}\|Xw - y\|_2^2 + \frac{\lambda}{2}\|w\|_2^2,
\end{equation}
and use the fitted coefficient vector $w$ as a candidate hyperplane normal for splitting.
We optimize this objective using Adam (learning rate $0.05$, batch size $2048$, base iterations $120$ with mild scaling at larger nodes).
This is an implementation convenience for fitting many node-local linear models without repeatedly forming and inverting large normal equations.

\subsection{Geometric Inductive Biases for Split Normals}
We incorporate three optional modules to shape $w$ before splitting.

\paragraph{Convolutional smoothing (\textsc{Conv}).}
We reshape $w$ into an $H\times W$ coefficient image aligned with the input grid, smooth it using a Gaussian kernel ($5\times5$, $\sigma=1.0$), and flatten back to a vector.
This encourages spatial coherence while preserving interpretability.

\paragraph{Importance pruning (\textsc{Prune}).}
We define a midpoint-aware importance score per feature,
\begin{equation}
\mathrm{imp}_i = |w_i| \cdot h_i,
\end{equation}
retain only the top-$k$ features (here $k=256$), and set the remaining coefficients to zero during tree growth (before evaluating the split).
This acts as a per-node compute budget.

\paragraph{Tilt dominance constraint (\textsc{Tilt}).}
A split direction can become ``spread'' across many pixels if non-dominant coefficients collectively rival the dominant one.
Let $k=\arg\max_i \mathrm{imp}_i$ be the most influential feature. We enforce the \emph{tilt inequality}
\begin{equation}
\sum_{i\neq k} |w_i|h_i \le \tau \, |w_k|h_k,
\end{equation}
with $\tau = 0.7$.
This inequality appears as a sufficient condition in the CMT analysis~\cite{CMT_original} for shrinking the node bounding box under geometric assumptions; here we use it as an inductive bias to encourage decisive, geometry-aligned splits.

\subsection{Deployment-Time Orientation Search (OS)}
At deployment time, OS adapts only the \emph{chosen input orientation}; model parameters remain fixed.
Given a discrete candidate set $\Phi=\{-40^\circ,-20^\circ,0^\circ,20^\circ,40^\circ\}$, OS selects an orientation $\phi$ by maximizing a normalized margin proxy aggregated across trees:
\begin{equation}
\phi^*(x)=\arg\max_{\phi\in\Phi} \sum_{t\in \mathcal{T}} \sum_{\ell \in \text{path}_t(x;\phi)}
\frac{\left| (x-m_\ell(\phi))^\top w_\ell(\phi)\right|}{\|w_\ell(\phi)\|_2 + \varepsilon},
\end{equation}
where $\mathcal{T}$ is the set of trees, $\text{path}_t(x;\phi)$ denotes the decision path of $x$ under rotation $\phi$ in tree $t$, and $\varepsilon$ is a small constant for numerical stability.
We then predict using the forest under $\phi^*(x)$.

To keep parameterization consistent across orientations, we rotate both inputs and coefficient grids using the same bilinear sampler: vectors such as $w_\ell$ and $m_\ell$ are treated as $H\times W$ coefficient images, rotated by $\phi$ with bilinear interpolation, and flattened back to vectors (similarly for the input image).

\section{Experimental Setup}
\subsection{Data and Evaluation Protocol}
\begin{figure}[t]
  \centering
  \includegraphics[width=0.55\linewidth]{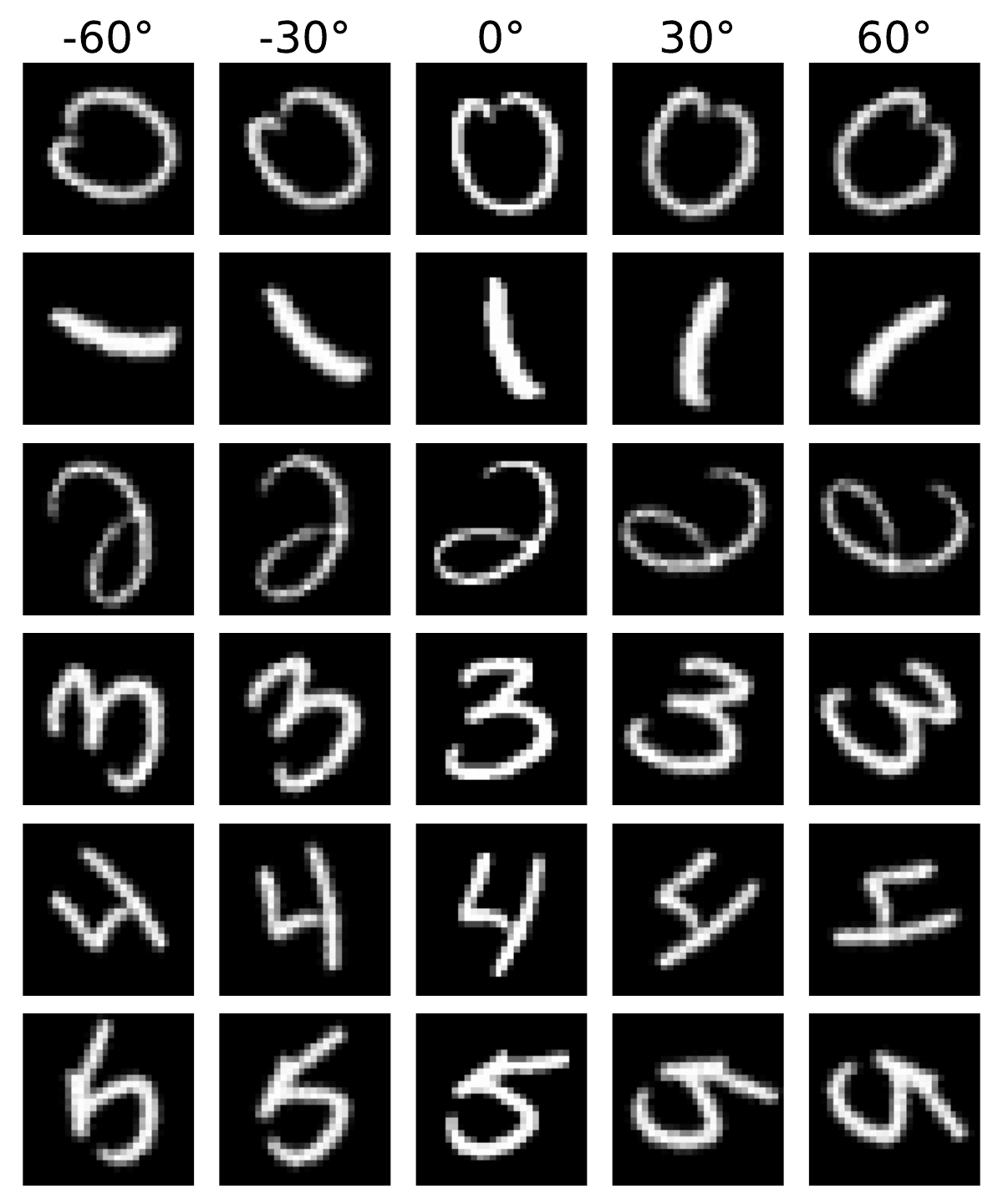}
  \caption{Rotated MNIST examples used in evaluation (bilinear resampling). We report results on the seven evaluation-time rotations $\Theta=\{-60^\circ,-40^\circ,-20^\circ,0^\circ,20^\circ,40^\circ,60^\circ\}$.}
  \label{fig:rotated_grid}
\end{figure}

We use full MNIST images (60,000 training and 10,000 test) at resolution $28\times 28$.
We evaluate robustness under seven evaluation-time input rotations
\[
\Theta=\{-60^\circ,-40^\circ,-20^\circ,0^\circ,20^\circ,40^\circ,60^\circ\},
\]
using bilinear interpolation for image resampling.
The perimeter target is computed on the unrotated images and reused for rotated inputs, consistent with rotation invariance.
To visualize the distortion induced by rotation and interpolation, Fig.~\ref{fig:rotated_grid} shows representative rotated digits.

\subsection{Models and Ablations}
All forests use 3 trees, bootstrap fraction $0.6$, maximum depth $6$, and minimum leaf size $200$.
Unless stated otherwise, we use \emph{hard gating} (HG) for routing.
For readability, we use short names for ablations; the mapping is summarized in Table~\ref{tab:model_key}.

\begin{table}[t]
\centering
\small
\caption{Model shorthand used throughout. HG denotes hard gating.}
\label{tab:model_key}
\setlength{\tabcolsep}{6pt}
\begin{tabular}{ll}
\toprule
\textbf{Short name} & \textbf{Description} \\
\midrule
Ridge & Single ridge regression on full training set \\
StdForest (HG) & Model-tree forest, no Conv/Tilt/Prune, HG \\
CMT-Conv & CMT with convolutional smoothing only \\
CMT-Conv+Tilt & CMT with Conv + Tilt \\
CMT-Conv+Prune & CMT with Conv + Prune \\
CMT-Full & CMT with Conv + Tilt + Prune \\
CMT-Full-HG & Same as CMT-Full but explicitly HG \\
\bottomrule
\end{tabular}
\end{table}

We enable OS on a subset of models: CMT-Conv, CMT-Conv+Tilt, and CMT-Full.
We report mean absolute error (MAE) for perimeter regression and accuracy for classification-as-regression.
All results correspond to a single run with random seed $42$.

\section{Results}
\subsection{Rotation Robustness on Perimeter Regression (Normal Prediction)}
\begin{figure}[t]
  \centering
  \includegraphics[width=0.9\linewidth]{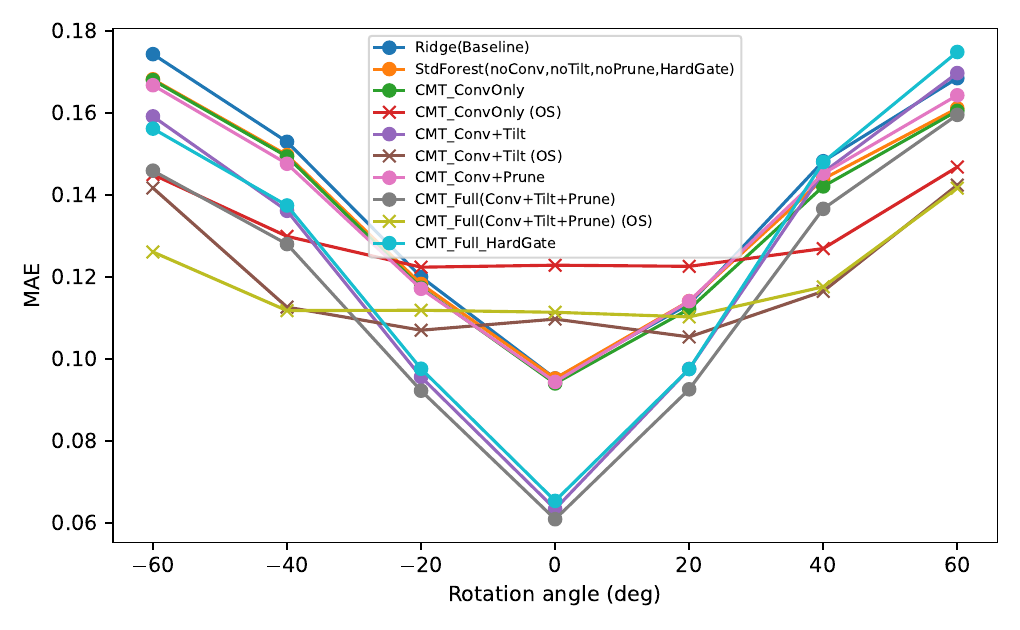}
  \caption{Perimeter regression MAE versus rotation angle. For models that support OS, we plot both normal prediction and OS-based prediction.}
  \label{fig:mae_vs_angle}
\end{figure}

We first evaluate robustness without OS (i.e., predicting using the evaluation rotation directly).
Table~\ref{tab:mae_normal} reports MAE across all rotations, and Fig.~\ref{fig:mae_vs_angle} visualizes the corresponding error curves.

The best non-adaptive model is \textbf{CMT-Full}, while CMT-Conv+Tilt is the strongest simpler geometry-aware ablation.
Both variants achieve their largest gains near the canonical $0^\circ$ setting, highlighting the benefit of decisive and spatially coherent split directions.

\begin{table}[t]
\centering
\small
\caption{Perimeter regression MAE under input rotations (normal prediction; no OS). Lower is better.}
\label{tab:mae_normal}
\setlength{\tabcolsep}{4pt}
\begin{tabular}{lcccccccc}
\toprule
Model & $-60^\circ$ & $-40^\circ$ & $-20^\circ$ & $0^\circ$ & $20^\circ$ & $40^\circ$ & $60^\circ$ & Avg \\
\midrule
Ridge & 0.1743 & 0.1529 & 0.1200 & 0.0952 & 0.1134 & 0.1482 & 0.1685 & 0.1389 \\
StdForest (HG) & 0.1682 & 0.1497 & 0.1183 & 0.0952 & 0.1141 & 0.1439 & 0.1611 & 0.1358 \\
CMT-Conv & 0.1680 & 0.1493 & 0.1173 & 0.0940 & 0.1121 & 0.1421 & 0.1605 & 0.1348 \\
CMT-Conv+Tilt & 0.1591 & 0.1361 & 0.0956 & 0.0633 & 0.0975 & 0.1455 & 0.1697 & 0.1238 \\
CMT-Conv+Prune & 0.1667 & 0.1476 & 0.1171 & 0.0944 & 0.1141 & 0.1452 & 0.1643 & 0.1356 \\
CMT-Full & \textbf{0.1459} & \textbf{0.1280} & \textbf{0.0922} & \textbf{0.0609} & \textbf{0.0926} & \textbf{0.1366} & \textbf{0.1595} & \textbf{0.1165} \\
CMT-Full-HG & 0.1562 & 0.1374 & 0.0976 & 0.0653 & 0.0975 & 0.1480 & 0.1748 & 0.1253 \\
\bottomrule
\end{tabular}
\end{table}

\subsection{Ablation Insights: Conv, Tilt, and Prune}
Table~\ref{tab:mae_normal} suggests three consistent patterns.
First, convolutional smoothing alone yields only modest improvement, indicating that spatial coherence by itself is insufficient.
Second, the tilt dominance constraint provides a large gain near the canonical orientation, consistent with the effect of enforcing decisive geometry-aligned splits.
Third, pruning acts primarily as a compute-control mechanism and becomes most effective when combined with tilt, yielding the best average MAE in this run.

\subsection{Deployment-Time Orientation Search (OS)}
We now evaluate OS for enabled models (Table~\ref{tab:mae_os}).
OS improves performance under severe rotations, but can be harmful near the canonical orientation, where the confidence proxy may spuriously favor nonzero rotations.
This failure mode suggests that practical deployment would benefit from priors favoring $\phi=0^\circ$ or rejection thresholds that only accept rotation changes when confidence gains are sufficiently large.

\begin{table}[t]
\centering
\small
\caption{Perimeter regression MAE with OS (enabled models only). Lower is better.}
\label{tab:mae_os}
\setlength{\tabcolsep}{4pt}
\begin{tabular}{lcccccccc}
\toprule
Model (OS) & $-60^\circ$ & $-40^\circ$ & $-20^\circ$ & $0^\circ$ & $20^\circ$ & $40^\circ$ & $60^\circ$ & Avg \\
\midrule
CMT-Conv + OS & 0.1449 & 0.1299 & 0.1224 & 0.1228 & 0.1226 & 0.1269 & 0.1468 & 0.1309 \\
CMT-Conv+Tilt + OS & 0.1417 & 0.1126 & 0.1070 & 0.1097 & 0.1053 & 0.1165 & 0.1424 & \textbf{0.1193} \\
CMT-Full + OS & 0.1261 & 0.1118 & 0.1118 & 0.1114 & 0.1102 & 0.1176 & 0.1416 & 0.1186 \\
\bottomrule
\end{tabular}
\end{table}

\subsection{Rotation Robustness on Classification-as-Regression}
\label{subsec:classification}
In addition to perimeter regression, we evaluate rotation robustness on digit recognition using the original MNIST labels.
To keep the learners unchanged, we implement multi-class classification via one-vs-rest regression:
we train 10 independent regression heads (one per class) with targets in $\{0,1\}$ and predict the class by $\arg\max_k s_k(x)$.

\begin{figure}[t]
  \centering
  \includegraphics[width=0.9\linewidth]{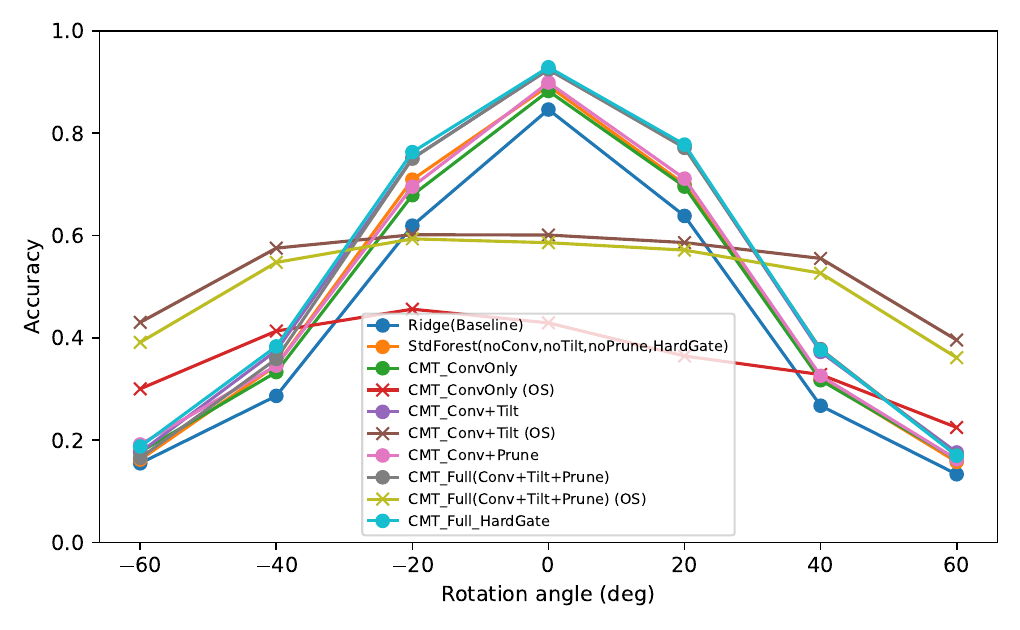}
  \caption{Classification-as-regression accuracy versus rotation angle. For models that support OS, we plot both normal prediction and OS-based prediction.}
  \label{fig:acc_vs_angle}
\end{figure}

Table~\ref{tab:acc_normal} reports accuracy under normal prediction (no OS), and Fig.~\ref{fig:acc_vs_angle} shows accuracy versus rotation.
All forest variants improve over Ridge across rotations, and the strongest variants maintain substantially higher accuracy under moderate rotations.

\begin{table}[t]
\centering
\small
\caption{Classification-as-regression accuracy under input rotations (normal prediction; no OS). Higher is better.}
\label{tab:acc_normal}
\setlength{\tabcolsep}{4pt}
\begin{tabular}{lcccccccc}
\toprule
Model & $-60^\circ$ & $-40^\circ$ & $-20^\circ$ & $0^\circ$ & $20^\circ$ & $40^\circ$ & $60^\circ$ & Avg \\
\midrule
Ridge & 0.1549 & 0.2866 & 0.6192 & 0.8461 & 0.6382 & 0.2674 & 0.1335 & 0.4208 \\
StdForest (HG) & 0.1621 & 0.3463 & 0.7092 & 0.8925 & 0.7000 & 0.3195 & 0.1577 & 0.4696 \\
CMT-Conv & 0.1771 & 0.3331 & 0.6787 & 0.8823 & 0.6955 & 0.3172 & 0.1629 & 0.4638 \\
CMT-Conv+Tilt & 0.1765 & 0.3752 & 0.7508 & 0.9242 & 0.7733 & 0.3725 & 0.1761 & 0.5069 \\
CMT-Conv+Prune & 0.1917 & 0.3457 & 0.6952 & 0.8989 & 0.7111 & 0.3259 & 0.1626 & 0.4759 \\
CMT-Full & 0.1654 & 0.3588 & 0.7503 & 0.9247 & 0.7712 & 0.3780 & 0.1709 & 0.5028 \\
CMT-Full-HG & \textbf{0.1878} & \textbf{0.3833} & \textbf{0.7629} & \textbf{0.9287} & \textbf{0.7776} & \textbf{0.3761} & \textbf{0.1698} & \textbf{0.5123} \\
\bottomrule
\end{tabular}
\end{table}

\paragraph{OS on classification.}
Table~\ref{tab:acc_os} reports accuracy with OS.
OS yields large gains under extreme rotations, but again degrades performance at $0^\circ$, mirroring the regression setting and reinforcing the need for calibration or priors in confidence-based angle selection.

\begin{table}[t]
\centering
\small
\caption{Classification-as-regression accuracy with OS (enabled models only). Higher is better.}
\label{tab:acc_os}
\setlength{\tabcolsep}{4pt}
\begin{tabular}{lcccccccc}
\toprule
Model (OS) & $-60^\circ$ & $-40^\circ$ & $-20^\circ$ & $0^\circ$ & $20^\circ$ & $40^\circ$ & $60^\circ$ & Avg \\
\midrule
CMT-Conv + OS & 0.3004 & 0.4133 & 0.4560 & 0.4292 & 0.3645 & 0.3282 & 0.2249 & 0.3595 \\
CMT-Conv+Tilt + OS & 0.4303 & 0.5753 & 0.6015 & 0.6009 & 0.5861 & 0.5552 & 0.3958 & \textbf{0.5350} \\
CMT-Full + OS & 0.3913 & 0.5474 & 0.5934 & 0.5859 & 0.5714 & 0.5264 & 0.3613 & 0.5110 \\
\bottomrule
\end{tabular}
\end{table}

\subsection{Compute Cost}
Training times are summarized in Table~\ref{tab:time}.
Tilt-enabled models are substantially slower in this implementation, likely due to increased split acceptance leading to more node-local fits.
This highlights a practical trade-off: stronger geometric inductive biases can improve robustness but may increase training cost.

\begin{table}[t]
\centering
\small
\caption{Training time (seconds) for each model in the reported run (seed=42).}
\label{tab:time}
\begin{tabular}{lc}
\toprule
Model & Train Time (s) \\
\midrule
Ridge & 0.74 \\
StdForest (HG) & 1.75 \\
CMT-Conv & 2.48 \\
CMT-Conv+Tilt & 43.90 \\
CMT-Conv+Prune & 1.72 \\
CMT-Full & 37.32 \\
CMT-Full-HG & 37.80 \\
\bottomrule
\end{tabular}
\end{table}

\section{Discussion}
\paragraph{Why does tilt help?}
The perimeter target depends on spatially coherent boundary structure.
The tilt inequality enforces dominance of a principal coordinate in the weighted sense $|w_i|h_i$, aligning with the geometric shrinkage intuition in the CMT analysis~\cite{CMT_original}.
Empirically, this reduces variance from aggregating many competing contributions and yields more decisive partitions, improving generalization near the canonical orientation.

\paragraph{When does OS help or hurt?}
Across both tasks, OS is most beneficial under severe rotations, where selecting a nearer canonical orientation can recover substantial performance.
However, OS can be harmful near $0^\circ$ when the confidence proxy spuriously favors nonzero rotations.
A practical remedy is to incorporate priors favoring $\phi=0^\circ$ or reject rotation changes unless the confidence gain exceeds a threshold.
More principled uncertainty estimates for angle selection are a natural direction for future work.

\paragraph{Full model interactions.}
In this run, combining pruning with tilt improves regression robustness, while OS provides limited additional benefit for the strongest base model due to canonical over-rotation.
This suggests that confidence-based orientation selection should be treated as a calibrated decision rule rather than a universally beneficial post-processing step.

\section{Limitations}
First, we evaluate on a synthetic regression label derived from MNIST rather than a real-world regression benchmark.
Second, OS uses a coarse angle grid and a simple confidence proxy; more principled uncertainty estimates may yield better orientation selection.
Third, node-local ridge models are optimized numerically; faster solvers or warm-starting could reduce training cost.

\section{Conclusion}
We presented a practical approach to rotation-robust regression using convolutional model trees and a deployment-time orientation search procedure.
Geometry-aware split biases improve robustness under rotations, and orientation search can further help under severe rotations without retraining.
However, confidence-based selection can be harmful near the canonical orientation, emphasizing the importance of calibration, priors, or rejection thresholds when applying deployment-time orientation selection in practice.

\section*{Reproducibility Checklist}
\begin{itemize}
  \item Dataset: MNIST images (60,000 train / 10,000 test), $28\times28$.
  \item Target: perimeter of thresholded mask ($\tau=0.1$), scaled by $\alpha=1/100$.
  \item Rotations evaluated: $\Theta=\{-60^\circ,-40^\circ,-20^\circ,0^\circ,20^\circ,40^\circ,60^\circ\}$.
  \item Orientation Search grid: $\Phi=\{-40^\circ,-20^\circ,0^\circ,20^\circ,40^\circ\}$.
  \item Forest: 3 trees, bootstrap 0.6, depth 6, min leaf 200, hard gating.
  \item Conv kernel: $5\times5$, $\sigma=1.0$; prune top-$k=256$; tilt $\tau=0.7$.
  \item Leaf ridge: Adam, lr 0.05, batch 2048, base iters 120, $L_2=0.1$.
  \item Random seed: 42.
\end{itemize}

\bibliographystyle{plainnat}
\bibliography{references}

\end{document}